\documentclass[10pt,twocolumn,letterpaper]{article}

\usepackage[pagenumbers]{cvpr} 

\usepackage[dvipsnames]{xcolor}

\definecolor{cvprblue}{rgb}{0.21,0.49,0.74}
\usepackage[pagebackref,breaklinks,colorlinks,citecolor=cvprblue]{hyperref}
\usepackage{asymptote}
\usepackage{MnSymbol}
\usepackage{tabularx}
\usepackage{multirow}
\usepackage{mathtools}

\renewcommand{\vec}[1]{\boldsymbol{#1}}
\DeclareMathOperator{\loss}{\mathcal{L}}

\def\paperID{15} 
\def\confName{CVPR}
\def\confYear{2024}

\title{Repeat and Concatenate: 2D to 3D Image Translation\\with 3D to 3D Generative Modeling}

\author{Abril Corona-Figueroa, Hubert P. H. Shum, Chris G. Willcocks\\
Department of Computer Science, Durham University, Durham, UK\\
{\small \url{https://github.com/abrilcf/3D-3D_repeat-concatenate}}
}

\begin{document}
\maketitle
\begin{abstract}
This paper investigates a 2D to 3D image translation method with a straightforward technique, enabling correlated 2D X-ray to 3D CT-like reconstruction. We observe that existing approaches, which integrate information across multiple 2D views in the latent space, lose valuable signal information during latent encoding. Instead, we simply repeat and concatenate the 2D views into higher-channel 3D volumes and approach the 3D reconstruction challenge as a straightforward 3D to 3D generative modeling problem, sidestepping several complex modeling issues. This method enables the reconstructed 3D volume to retain valuable information from the 2D inputs, which are passed between channel states in a Swin UNETR backbone. Our approach applies neural optimal transport, which is fast and stable to train, effectively integrating signal information across multiple views without the requirement for precise alignment; it produces non-collapsed reconstructions that are highly faithful to the 2D views, even after limited training. We demonstrate correlated results, both qualitatively and quantitatively, having trained our model on a single dataset and evaluated its generalization ability across six datasets, including out-of-distribution samples.
\end{abstract} 
\section{Introduction}

2D to 3D image translation is a class of computer vision problems where the goal is to learn the mapping between one or more 2D images and a corresponding 3D volumetric image. Years of research in this topic have given rise to many image and graphics applications, such as augmented reality \cite{feng2023enhancing:ar1,liu2024pa:ar2}, sensor fusion \cite{oh20232:sf1,lee20232:sf2}, scene rendering \cite{corona2022mednerf:7, jin2023semi:rend}, and multimodal translation \cite{isaac2022cross, sasaki2021unit, zhan2023multimodal}. The latter has attracted attention in the medical domain, including 2D X-ray, Computed Tomography (CT), Magnetic Resonance Imaging (MRI), Ultrasound, among others, due to the potential to translate from cheap, low-quality and available medical imaging modalities to those that are more expensive, with high-waiting times, or exhibit harmful ionizing radiation. While translating between image modalities of common dimensionality (i.e., 2D to 2D, 3D to 3D) has achieved notable results \cite{zhan2023multimodal} due to its seemingly straightforward adaptation of specific state-of-the-art (SOTA) deep learning models, translating between data of distinct dimensionality poses further challenges. In particular, converting an image from lower-dimension to a higher-dimensional representation (e.g., 2D to 3D) is considered a reconstructive generative modelling problem \cite{wang2020deep, bond2021deep}.

\begin{figure}[t]
    \centering
    \includegraphics[width=1.0\linewidth]{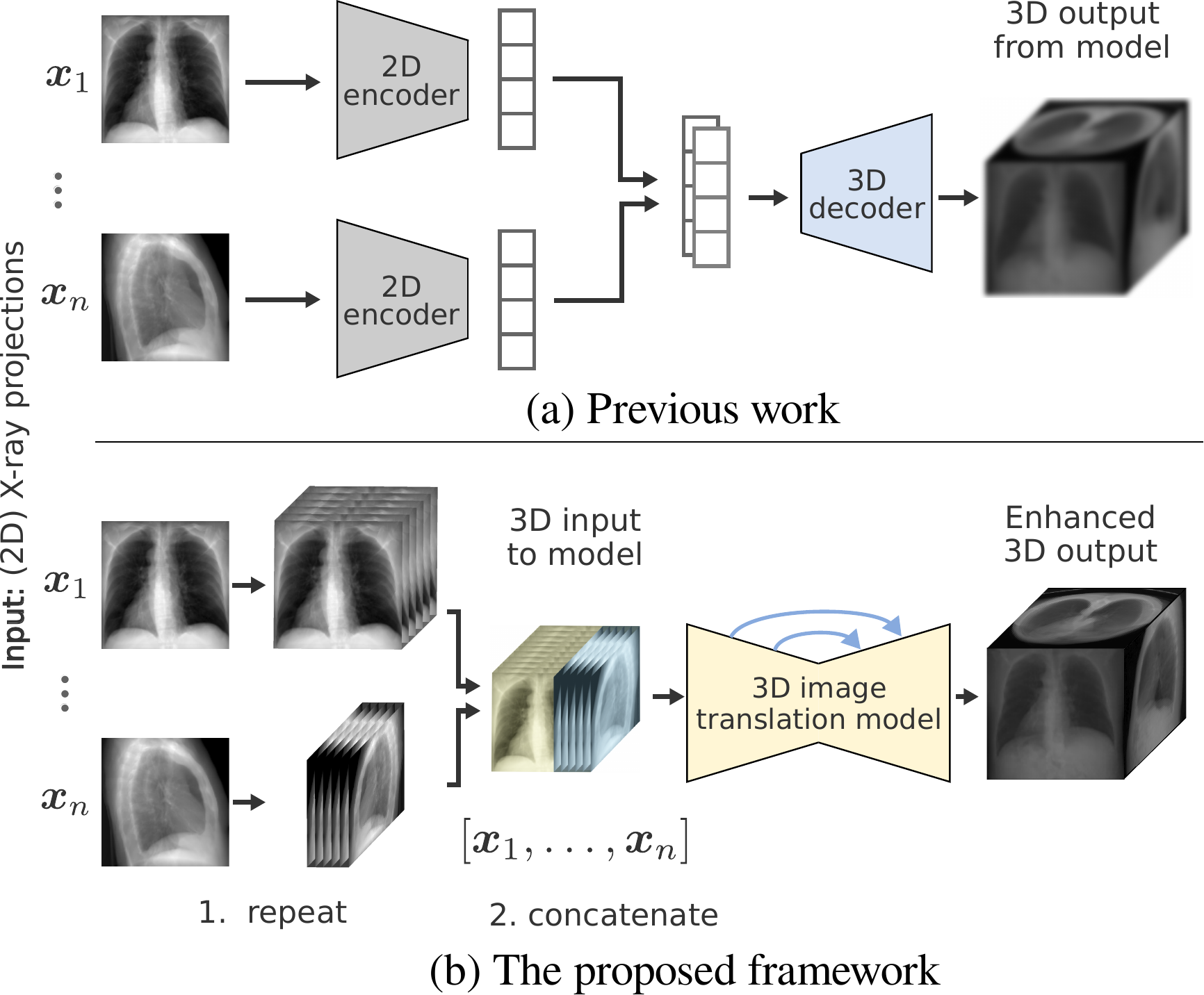}
    \caption{(a) Previous approaches focus on 2D to 3D mapping, often employing asymmetric architectures and compressed latent encoding. (b) In contrast, we propose 3D to 3D mapping from repeated and concatenated inputs, enabling faster training with highly correlated outputs without latent compression, even with small datasets (a few hundred images).}
    \label{fig:prepro_comp}
    \vspace{-10pt}
\end{figure}

One highly relevant application in the medical field involves obtaining 3D CT representations from 2D X-ray projections \cite{maken20232d}. However, unlike image super-resolution approaches, bridging the 2D to 3D data dimensionality gap presents a unique modeling challenge to estimate spatial missing details \cite{wang2020deep}. Moreover, machine learning models achieve remarkable results thanks to abundant natural image data, but medical models often struggle due to the limited size of medical datasets. Furthermore, real-world medical 3D datasets involve volumetric features with varying density structures, making hallucinated data a significant concern that can prove counterproductive \cite{paavilainen2021bridging}.

Existing methods approach 2D to 3D medical image translation through asymmetrical architectures and/or incorporate various regularization techniques to enforce plausible reconstructions \cite{ying2019x2ct:8, ratul2021ccx} (Fig. \ref{fig:prepro_comp}). While these techniques may generate high-quality outputs, even with high-frequency details, they often lack correlation to the input data. This means that the model becomes overly reliant on the latent prior, potentially ignoring the original 2D signal. The situation is further exacerbated if the training data is limited (few hundred images), so when the model is presented with inputs significantly different from the training dataset, such as out-of-distribution samples, it is likely to perform poorly making it unusable for practical scenarios.

\begin{figure*}[ht]
    \centering
    \includegraphics[width=1.0\linewidth]{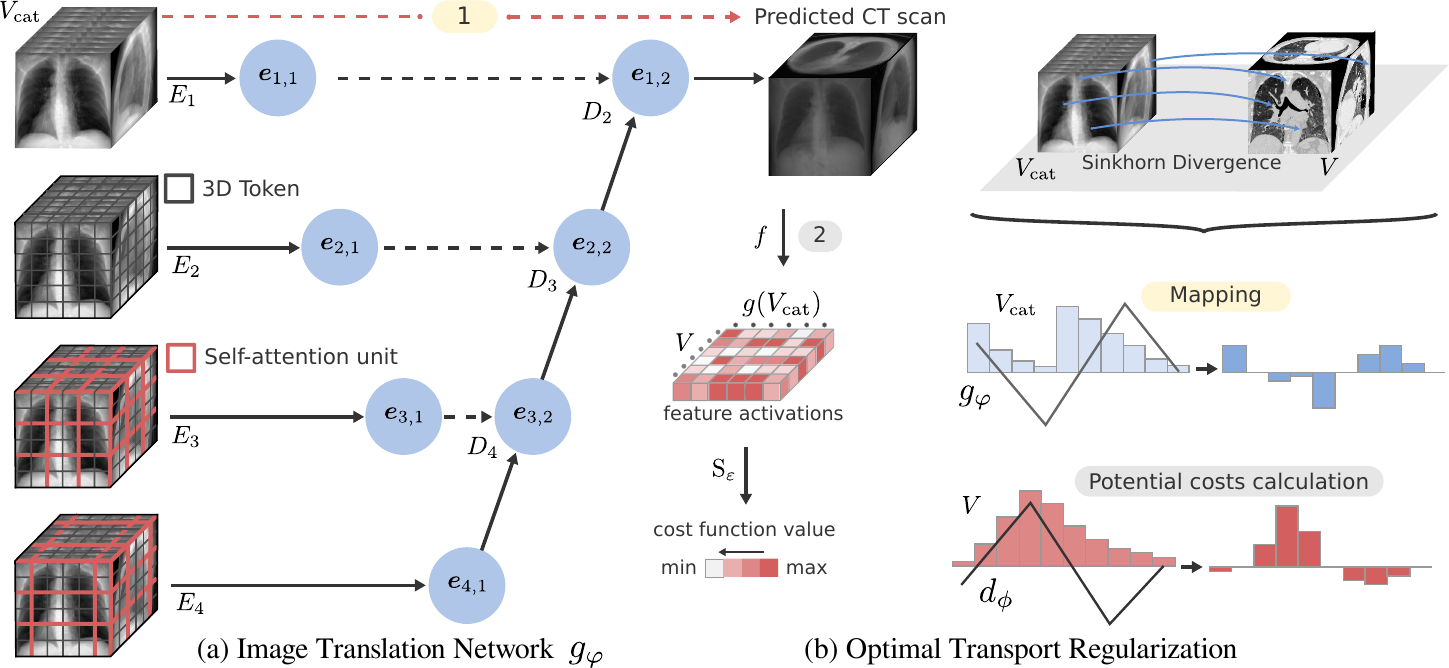}
    \caption{Proposed 2D to 3D image translation approach. (a) We learn the mapping between 2D inputs and their corresponding 3D representation adapting a Swin UNe-Transformer $g_\varphi$ \cite{hatamizadeh2021:swin} for image translation. (b) Optimization is based on the dual optimal transport regularization between networks $g_\varphi$ and $d_\phi$, comparing data points from the two image spaces using the Sinkhorn divergence $\mathrm{S}_\mathrm{\varepsilon}$ on the activations of a feature extractor $f$.}
    \label{fig:method}
\end{figure*}

In this paper, we propose a simple 2D to 3D mapping translation framework that addresses the correlation issue, ensuring highly-associated outputs even in limited datasets. We achieve this by a preprocessing step that preserves 2D information content throughout the network transformations without relying on other priors. First, we repeat the various 2D input projections to match the output 3D target depth (Fig. \ref{fig:prepro_comp}). These 3D volumes are then concatenated into a single higher-channel 3D volume. Then we propose a 3D to 3D conditional generative modeling approach that applies neural optimal transport with the de-biased Sinkhorn divergence. Such an approach effectively allows mass splitting; we find this particularly suitable for our task, as the original 2D inputs contain valuable information content, requiring each area of the 2D inputs to contribute to entire 3D regions of details in the synthesized 3D image (Fig. \ref{fig:method}).

The approach was found to give significant improvement in terms of generalization, while being stable to train with only a few hundred images (ideal for medical/real-world datasets). Our evaluation showed that the method generates plausible reconstructions after only 2,000 training optimization steps, which can generalize to out-of-distribution inputs after approximately 28 hours of training.

To summarize, our contributions are
\begin{enumerate}
    \item An alternative and simple 2D to 3D image translation framework that effectively reconstructs a CT volumetric representation given only one or any number of X-ray projections, with potential application in clinical use (e.g., cutting down costs and radiation to patients).
    \item A processing pipeline that retains all 2D information throughout the 3D 
    transformation network without loosing information content in the latent encoding.
    \item Demonstrable generalization with very limited data (a few hundred images) trained in less than 28 hours. The processing step and approach can be directly applied to other generative modelling approaches and applications.
\end{enumerate}
\section{Related Work}

\paragraph{2D to 3D image translation:} Several methods have attempted to obtain 3D representations from 2D images for medical applications, such as CT reconstruction from X-rays. Conditional approaches, generating 3D images from one or two 2D images are often based on Generative Adversarial Networks (GANs) \cite{goodfellow2020generative} and Variational Autoencoders (VAEs) \cite{kingma2013auto}. These methods transform noise sampled from a Gaussian prior, which is concatenated with a latent representation of the 2D data; this may loose or hallucinate important 2D signal information not captured in the latent encoding. Strategies to mitigate this include incorporating 3D priors from real data \cite{9698875:9} or simply generating 2D slices as output, which are stacked into a 3D representation \cite{9463166:10}.

Another line of work involves asymmetric autoencoder architectures, where a 2D encoder extracts spatial features, which are then expanded to fit into a 3D decoder \cite{nashed2021end:1,kyung2023perspective:2,stojanovski2022efficient:3,gunduzalp20213d:4,LIU2023:6}. To mitigate information loss at the bottleneck, additional objectives are learned, enforcing a more interpretable latent space or adding an additional 2D encoder \cite{ying2019x2ct:8}. Recent approaches explore implicit neural representations \cite{corona2022mednerf:7,maas2023nerf,song2023piner} or two-stage vector-quantized approaches combined with diffusion \cite{Corona-Figueroa_2023_ICCV:5}. In contrast, we explore a simpler approach that generalizes from small datasets ($\sim$1k images) in a 3D to 3D setting without relying on more complex asymmetric architectures.
\vspace{-2pt}
\paragraph{Regularization and prior knowledge:} Generalizing beyond the examples in the training set is a fundamental aspect of machine learning models for medical applications. Key problems are mode collapse and overfitting, which arise due to the limited availability and real-world complexity of medical datasets. This is further exacerbated by data noise and heterogeneity among different imaging modalities. Fortunately, regularization techniques such as parameter constraints or augmentation can mitigate overfitting \cite{moradi2020survey,tian2022comprehensive} and help alleviate data shortages.

Data augmentation, in particular, is an established solution to mitigate overfitting \cite{kebaili2023deep, garcea2023data} by introducing semantic-preserving data transformations such as rotation, translation, and contrast adjustments. However, selecting and combining augmentations is a delicate process, especially to prevent undesirable outcomes such as the model learning to generate the augmented distribution and deviating from its original objective \cite{karras2020training}. In this context, incorporating prior knowledge from initial assumptions about the data distribution can guide learning for improved generalization, such as additional conditional information in generative modeling or via additional terms to the main objective functions \cite{weinberger2020learning, ulyanov2018deep, fortuin2022priors}. In our approach, we know that 2D X-rays capture the underlying 3D objects over a range of depths, acting as a prior that motivates our repeat and concatenate mapping strategy. We find this facilitates improved generalization for both in- and out-of-distribution inputs, even when trained on small datasets.
\vspace{-5pt}
\paragraph{Deep generative modeling:} 2D to 3D image translation involves modeling the probability of unobserved 3D regions which is therefore a generative modeling problem. Deep generative modeling (DGM) is a very large field, from which mainstream approaches include GANs, VAEs, normalizing flows, autoregressive models and probablistic diffusion models~\cite{bond2021deep}. These methods classically balance an empirically observed trilemma of modeling quality, mode coverage and the number of `steps' taken during sampling~\cite{xiao2022tackling}. VAEs and normalizing flows are well-known for lower quality sampling, while GANs suffer from mode collapse, capturing only part of the distribution during training. Various strategies, such as weight clipping, gradient penalties, and spectral normalization, have been proposed to mitigate these issues, but they often fail to address fundamental challenges rooted in optimal transport (OT) theory.
\paragraph{Neural optimal transport:} In contrast, neural optimal transport offers a more nuanced approach for mapping probability distributions, which is especially relevant for handling the intricacies of medical imaging data. Traditional methods like $f$-divergences, commonly employed in GANs, perform poorly with smaller, high-dimensional datasets, failing to provide meaningful gradients for effective training \cite{tolstikhin2018wasserstein}. Capturing variability in limited medical data is crucial, and leveraging regularity in human datasets is one approach. For example, organs and bones consistently occupy locations in the human chest, allowing neural networks to learn the mapping between X-ray and 3D CT volumes. OT is relevant in this setting by providing a rigorous distance metric that respects the geometric properties of the distributions. It computes the minimal cost required to transform one distribution into another, offering a coherent and robust measure of similarity that is both intuitive and stable~\cite{feydy2020geometric}. This geometrically-aware approach enables a more direct and meaningful comparison of medical images across different modalities, potentially facilitating the development of more accurate and reliable diagnostic tools.

\section{Method} \label{s:method}
We initially detail our 2D to 3D processing approach (Fig. \ref{fig:prepro_comp}), and then discuss our 3D to 3D generative modeling regularized with optimal transport (Fig. \ref{fig:method}). We also justify how the 3D to 3D mapping approach improves correlation by analyzing the information content through the transformation strategies.

\subsection{Processing pipeline}


At a high-level, we wish to investigate whether multi-view 2D information can be combined for accurate 3D synthesis without latent encoding. Previous approaches integrate multi-view information in the latent space, as it typically captures a degree of spatial and rotational invariance. However the latents, even with information-rich modeling, fail to retain the full signal from the input 2D views. While, through modern deep generative modeling, they may synthesize high-quality signals with high-frequency details, these are typically prone to over-hallucination~\cite{Corona-Figueroa_2023_ICCV:5} which can potentially pose fatal in real-world medical application. In this application, we would rather have blurriness and uncertainty at the expense of outputs that are highly-correlated to their corresponding 2D inputs.

To achieve this objective, we propose increasing the dimensionality at the 2D inputs to ensure the signal can influence an entire region of the 3D volume, and concatenating these inputs ensuring that information content loss is reduced within 3D to 3D residual architectures.



More formally, we outline our processing pipeline for $N$ input X-ray projections. Let $I \in \mathbb{R}^{C \times H \times W}$ be an input 2D X-ray, representing a single view with height $H$, width $W$ and channels  $C$. For a set of $N$ views, we have $\{I^i\}_{i=1}^N$, each aiming to contribute to the reconstruction of a 3D CT volume $V \in \mathbb{R}^{C \times H \times W \times D}$, where $D$ is the depth. 

We `stretch' the 2D inputs to match the dimensions of the target 3D volume; we repeat each view $D$ times by $\Gamma : \mathbb{R}^{C \times H \times W} \to \mathbb{R}^{C \times H \times W \times D}$ and coarsely align them by transposing views that differ by 90-degrees (leaving the others unchanged, see our experiments section on \textit{Views alignment} for further discussion on this). This is applied for each of the $N$ views, which are concatenated across the depth axis (denoted by $\bigoplus$) yielding a composite volume $V_{\text{cat}} \in \mathbb{R}^{NC \times H \times W \times D}$, where
\begin{equation}
    V_{\text{cat}} = \bigoplus_{i=1}^N \Gamma(I^i),
\end{equation}
and the reconstructed 3D volume is modeled by a U-Net based architecture $g_{\varphi}^{\text{unet}} : \mathbb{R}^{NC \times H \times W \times D} \to \mathbb{R}^{C \times H \times W \times D}$.

\subsection{Mapping network and generative modeling}

For our 3D to 3D mapping network $g_\varphi$ with parameters $\varphi$ we considered a residual U-Net \cite{ronneberger2015:unet} equipped with self-attention following the Swin
UNEt-TRansformer (Swin UNETR) model \cite{hatamizadeh2021:swin}, applied for image translation instead of image segmentation. Based on experimenting with a variety of off-the-shelf U-Net backbones, we settled on Swin UNETR as it empirically generated higher image quality outputs across all metrics. 

\subsubsection{Modeling with neural optimal transport}

For the generative modeling task, we applied the dual regularized optimal transport ($\mathrm{OT}$) learning approach \cite{gozlan2017kantorovich} using the geometric de-biased Sinkhorn divergence ${\mathrm{S}}_\varepsilon(\cdot)$ \cite{feydy2019interpolating} as the cost function. The Sinkhorn divergence is suitable for this task as it is an efficient yet positive and definite approximation of OT that interpolates between Maximum Mean Discrepancies (MMD) and OT. Specifically, we incorporate a feature extractor $f$ that reduces the dimensionality of $g(V_{\text{cat}})$ and $V$ through a mapping into a reduced dimensional vector as suggested in \cite{genevay2018learning}. We parametrized $f$ as a neural network and calculate the cost function in the latent space,
\begin{equation}
    {\mathrm{S}}_\mathrm{\varepsilon}(\alpha, \beta) \vcentcolon  = \mathrm{OT}_\varepsilon(\alpha, \beta) - \frac{1}{2}\mathrm{OT}_\varepsilon(\alpha, \alpha) - \frac{1}{2}\mathrm{OT}_\varepsilon(\beta, \beta),
\end{equation}
where $\alpha$ represents the distribution of predicted features and $\beta$ the distribution of ground truth 3D CT features extracted from $f$. While calculating the cost in the data space might suffice in some cases, we find that incorporating the feature extractor $f$ is beneficial with 3D data as it reduces its dimensionality and makes our mapping network less susceptible to mode collapse.

This modeling approach is similar to the min-max objective in GANs where the Swin UNETR mapping network $g_\varphi$ (Fig. \ref{fig:method}a) replaces the generator and a residual discriminator $d_\phi$ assigns transportation costs to the produced samples $g_\varphi(V_\text{cat})$. We trained our networks using:
\begin{equation}
    \loss_g = \mathbb{E}_{(V_{\text{cat}},V)\sim p_{\text{data}}}\left[ {\mathrm{S}}_\mathrm{\varepsilon} (g(V_{\text{cat}}), V)- \lambda \,d(g(V_{\text{cat}}))\right],
\end{equation}
\begin{equation}
    \loss_d = \mathbb{E}_{(V_{\text{cat}},V)\sim p_{\text{data}}} \left[d(g(V_{\text{cat}})) - d(V)\right].
\end{equation}
The optimization of the network $d$ involves finding a function that yields the minimal cost associated with transporting mass from each point in $V_\text{cat}$ to each point in $V$. Overall this allows us to align the distributions of features from ground truth CT samples with our samples from $g_\varphi$ attained from the concatenated volumes from X-ray views $V_{\text{cat}}$, through regularized dual OT (Fig. \ref{fig:method}b). 

Even though we can use the L2 distance to approximate the OT plan, it does not capture the underlying structure of the distributions or how to transport mass from one to another. In contrast, the Sinkhorn divergence induces a scaling process that ensures, at each step, the resulting transportation plan satisfies probability distribution constraints, thus helps avoiding common training instability issues in adversarial training. Furthermore, our model relies solely on 2D views, promoting deterministic mapping and reducing hallucination by avoiding sampling from additional distributions, such as Gaussian, typical of multimodal image translation. We find that OT regularization greatly reduces overfitting but may introduce blur in the generated outputs due to the unconstrained nature of 2D-3D translation.

\subsection{Justification}

We justify our processing pipeline by examining information loss through different approaches. In previous works, encoded features $\mathbf{z} \in \mathbb{R}^M$ are typically obtained through an encoding function $\mathbf{z} = f^{\text{enc}}(I)$, where typically, $M < CHW$, indicating a reduction in dimensionality. The decoding function $V = f^{\text{dec}}(\mathbf{z})$ similarly tries to reconstruct a volume given the latent. 

\paragraph{Information content \(\textrm{I}(\cdot)\)}
Let \(\textrm{I}(\cdot)\) quantify the informational content of an image or volume, indicating that \(\textrm{I}(V)\) is maximized when \(V\) contains the full detail and structure inherent to the original 3D object.

We show that the repetition and concatenation approach retains information when integrated via U-Net over multiple views, compared to classicial encoder-decoder approaches
\[g^{\text{unet}}\left( V_{\text{cat}}\right) \succ f^{\text{dec}}\left(\oplus_{i=1}^N f^{\text{enc}}(I^i)\right),\]
where \(\succ\) denotes higher fidelity (closeness to original) in 3D reconstruction, directly correlated with informational content \(\textrm{I}(\cdot)\).

\paragraph{Information loss in encoding and decoding}
Dimensionality reduction through encoding implies: $M < CHW,$
highlighting a reduction in the capacity to represent the full informational content of the original volume. This reduction leads to inherent information loss, which the decoding process cannot fully recover as
\begin{equation}
    M < CHW \implies \textrm{I}(f^{\text{dec}}(\mathbf{z})) < \textrm{I}(I),
\end{equation}
due to the lossy nature of compression in \(f^{\text{enc}}\) and the limited ability of \(f^{\text{dec}}\) to fully recover original information.

\paragraph{Information retention through skip-connections}
The skip connections in \(g^{\text{unet}}\) enable direct transfer of features across layers, preserving and refining informational content where
\begin{equation}
    \textrm{I}(g^{\text{unet}}(V_{\text{cat}})) \approx \textrm{I}(V_{\text{cat}}) = \textrm{I}(I),
\end{equation}
especially when \(V\) is constructed from repeating \(I\) across the depth dimension (without information loss), thereby maintaining high fidelity from input to output.

\paragraph{Information retention across views}
The concatenation of repeated views tends to preserve informational content, while the concatenation of encoded views may lead to information loss due to aggregation
\begin{equation}
    \textrm{I}\left(\oplus_{i=1}^{N} \Gamma(\vec{x}^i)\right) \approx \sum_{i=1}^N \textrm{I}(\Gamma(I^i)),
\end{equation}
\begin{equation}
    \textrm{I}\left(\oplus_{i=1}^N g^{\text{enc}}(\vec{x}^i)\right) \leq \sum_{i=1}^N \textrm{I}(I^i).
\end{equation}

Therefore, the combination of multiple views through concatenation and subsequent processing with \(g^{\text{unet}}\) tends to retain information content, whereas concatenating the encodings leads to significant loss in the correlation on aggregate, leading to
\begin{equation}
    \textrm{I}\left(g^{\text{unet}}\left(\oplus_{i=1}^N \Gamma(I^i)\right)\right) > \textrm{I}\left(f^{\text{dec}}\left(\oplus_{i=1}^N f^{\text{enc}}(I^i)\right)\right),
\end{equation}
indicating the potential of enhanced fidelity of 3D reconstruction achieved through a U-Net based architecture with repeated concatenation and skip connections over latent integration frameworks.
\begin{figure}[t]
    \centering
    \begin{subfigure}[b]{\columnwidth}
        \centering
        \fontsize{8pt}{9pt}\selectfont
        \begin{tabularx}{\linewidth}{l *{4}{>{\centering\arraybackslash}X}}
            \multicolumn{5}{c}{LIDC-IDRI dataset \cite{armato2011:lidc} (in-of-distribution inputs) using two views} \\
            \midrule[0.1pt]
            Experiment         & $\uparrow$ SSIM   & $\uparrow$ PSNR   & $\downarrow$ MSE   & $\downarrow$ MAE  \\
            \midrule[0.1pt] 
            2D-3D AE            & 0.1545             & 3.804              & 995.2582            & 5.81              \\
            2D-3D AE + L2 Norm. & 0.2086             & 10.833             & 197.2649            & 2.5103            \\
            3D-3D U-Net         & \bf{0.4787}        & \bf{22.063}        & \bf{58.7546}        & \bf{0.9264}       \\
        \end{tabularx}
        \vspace{0pt}
    \end{subfigure}
    \begin{subfigure}[b]{\columnwidth}
        \centering
        \includegraphics[width=1\linewidth]{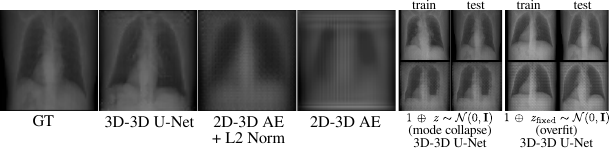} 
    \end{subfigure}
    \caption{Effect of reformulating 2D-3D mapping into 3D-3D.}
    \label{fig:baseline_experiments}
\end{figure}
\begin{figure*}[t]
    \centering
    \includegraphics[width=1.0\linewidth]{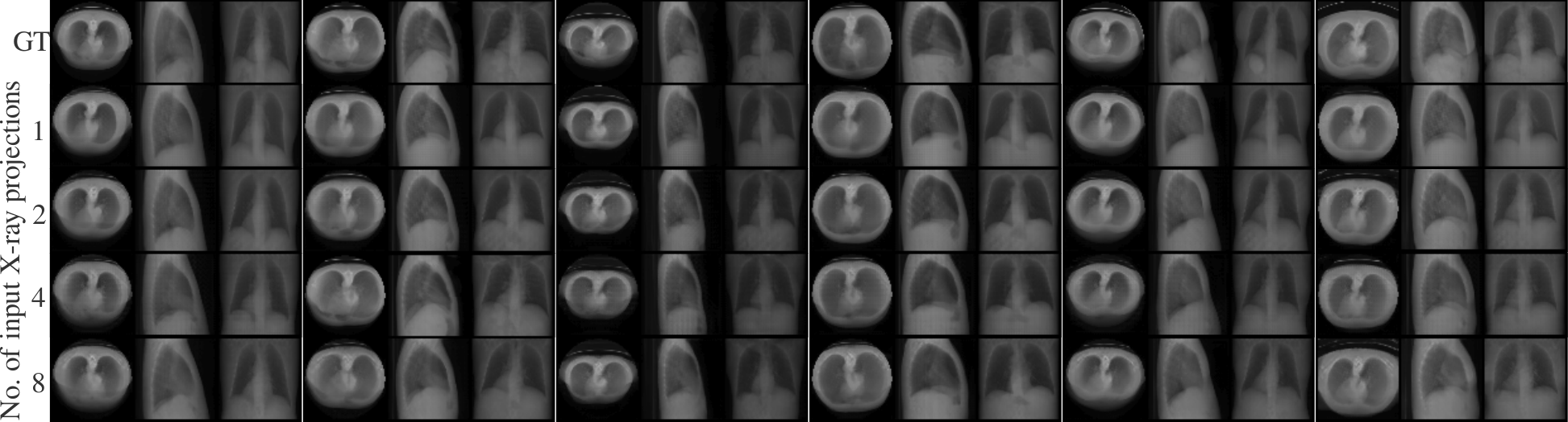}
    \caption{Example projections from generated 3D CT volumes from inputs using one, two, four and eight X-rays; obtained from our 3D-3D translation approach with Swin UNETR backbone. We use testing instances from LIDC-IDRI dataset \cite{armato2011:lidc}.}
    \label{fig:views_ablation}
    \vspace{-5pt}
\end{figure*}

\section{Experiments} \label{s:experiments}

We trained our models on the LIDC chest dataset \cite{armato2011:lidc}, which consists of only 916 training CT scans, and tested them on both in-distribution and out-of-distribution inputs from six lung datasets. We created paired datasets generating digitally reconstructed radiographs as 2D inputs from the CT scans using the open-source software Plastimatch, following previous work \cite{ying2019x2ct:8, ratul2021ccx, corona2022mednerf:7, Corona-Figueroa_2023_ICCV:5}. Our model's convergence plateaued after 2,000 iterations, taking only an average of 28 hours to reach 5,000 steps (selected weight). The training runs were performed on NVIDIA TITAN RTX with a batch size of eight using PyTorch.

\begin{table}
\begin{center}
\fontsize{8pt}{9pt}\selectfont
    \begin{tabular}{ccccc}
    \multicolumn{5}{c}{In-distribution inputs (LIDC-IDRI dataset \cite{armato2011:lidc})}
            \\ \midrule[0.1pt]
No. input views &  $\uparrow$  SSIM  &  $\uparrow$ PSNR  &  $\downarrow$ MSE  &  $\downarrow$ MAE    
    	\\    \midrule[0.1pt] 
1 $\oplus\,\, z\sim\mathcal{N}(0,\boldsymbol{\mathrm{I}})$ &  0.2337  &  20.2827 &  0.0403  &  0.1342
             \\
1  &  0.4891  &  23.2198 &  0.0214  &  0.0751 
             \\
 2  &  \underline{0.5272}  & 24.3192 &  0.0180 &  \underline{0.0665} 
             \\
 4  & 0.5129  &  \underline{24.3358} &  \underline{0.0167} &  0.0666 
             \\
 8  & \bf{0.5402} & \bf{24.6352} &  \bf{0.0155} &  \bf{0.0613}                            
             \\
    \\
    \multicolumn{5}{c}{Out-of-distribution inputs (MIDRC-1b dataset \cite{tsai2020:midrc})}
            \\ \midrule[0.1pt]
1 $\oplus\,\, z\sim\mathcal{N}(0,\boldsymbol{\mathrm{I}})$ &  0.1353  &  15.3223 &  0.0800  &  0.2334
             \\
1  & \underline{0.4043}    & 20.0800  &                                         0.0526    &  0.1510
             \\
 2  &  \bf{0.4048}      & \underline{20.3636}    &  \underline{0.0505}    &  \underline{0.1474}
             \\
 4  &  0.3779     & \bf{21.2115}    &  \bf{0.0439}     &  \bf{0.1337}
             \\
 8  &  0.2029      & 17.7972    &    0.0777     &  0.2029                            
             \\
    \end{tabular}
\end{center}
\vspace{-10pt}
\caption{Primary results of our mapping approach with varying input views. Each model was trained on LIDC dataset \cite{armato2011:lidc} and tested for both in- and out-of-distribution inputs from MIDRC-1b dataset \cite{tsai2020:midrc}. We computed metrics five times with different random seeds and report their average.}
\label{tab:views_ablations}
\vspace{-10pt}
\end{table}

\begin{table}
\begin{center}
\fontsize{8pt}{9pt}\selectfont
\resizebox{\columnwidth}{!}{%
    \begin{tabular}{llcccc}
    \multicolumn{6}{c}{\fontsize{10pt}{12pt}\selectfont (a) In-of-distribution inputs}
            \\ \midrule[0.1pt]
\bf{Dataset} &  \bf{Method} & $\uparrow$  SSIM  &  $\uparrow$ PSNR  &  $\downarrow$ MSE  &  $\downarrow$ MAE    
    	\\    \midrule[0.1pt] 
\multirow{3}{*}{LIDC-IDRI \cite{armato2011:lidc}} &  X2CT-GAN \cite{ying2019x2ct:8} &  0.321 &  19.68 &  0.045  &  0.151
             \\ 
& CCX-rayNet \cite{ratul2021ccx} &  \underline{0.386} &  \underline{22.66} &  \underline{0.032}  &  \underline{0.108}
             \\
& Ours  &  \bf{0.527}  &  \bf{24.35} &  \bf{0.018}  &  \bf{0.066} 
             \\
    \\
    \multicolumn{6}{c}{\fontsize{10pt}{12pt}\selectfont (b) Out-of-distribution inputs}
            \\ \midrule[0.1pt]
\multirow{3}{*}{COVID-19-NY-SBU \cite{saltz2021stony:covid-19}} & X2CT-GAN \cite{ying2019x2ct:8}                                       &  \underline{0.236} &  16.74 &  0.089  &  0.199
             \\ 
& CCX-rayNet \cite{ratul2021ccx} &  0.205 &  \underline{19.03} &  \underline{0.054}  &  \underline{0.144}
             \\
& Ours  &  \bf{0.400}  &  \bf{22.78} &  \bf{0.022}  &  \bf{0.077} 
             \\  \midrule[0.1pt]
\multirow{3}{*}{SPIE-AAPM-NCI \cite{armato2015:spie}} & X2CT-GAN \cite{ying2019x2ct:8}                                       &  \underline{0.174} &  15.63 &  0.115  &  0.245
             \\
& CCX-rayNet \cite{ratul2021ccx} &  0.130 &  \underline{17.87} &  \underline{0.080}  &  \underline{0.196}
             \\
& Ours  &  \bf{0.399}  &  \bf{22.04} &  \bf{0.026}  & \bf{0.087} 
             \\  \midrule[0.1pt] 
\multirow{3}{*}{MIDRC-1b \cite{tsai2020:midrc}} & X2CT-GAN \cite{ying2019x2ct:8}                                       &  \underline{0.220} &  14.70 &  0.156  &  0.360
             \\
& CCX-rayNet \cite{ratul2021ccx} &  0.114 &  \underline{18.06} &  \underline{0.076}  &  \underline{0.228}
             \\
& Ours  &  \bf{0.404}  &  \bf{20.36}  &  \bf{0.050}  &  \bf{0.147}  
             \\ \midrule[0.1pt] 
\multirow{3}{*}{ANTI-PD-1 \cite{madhavi2019data:anti-pd}} & X2CT-GAN \cite{ying2019x2ct:8}                                       &  \underline{0.286} &  18.04 &  0.072  &  \underline{0.164}
             \\
& CCX-rayNet \cite{ratul2021ccx} &  0.205 &  \underline{18.15} &  \underline{0.067}  &  0.167
             \\
& Ours  &  \bf{0.349}  &  \bf{20.93} &  \bf{0.040}  &  \bf{0.112} 
             \\ \midrule[0.1pt] 
\multirow{3}{*}{LCTSC \cite{yang2017data:lctsc}} & X2CT-GAN \cite{ying2019x2ct:8}                                       &  \underline{0.326} &  19.20 &  \underline{0.052}  &  \underline{0.125}
             \\
& CCX-rayNet \cite{ratul2021ccx} &  0.206 &  \underline{19.75} &  0.062  &  0.156
             \\
& Ours  &  \bf{0.331}  &  \bf{20.38} &  \bf{0.040}  &  \bf{0.109} 
             \\ \midrule[0.1pt] 
\multirow{3}{*}{NSCLC \cite{bakr2018radiogenomic:nsclc}} & X2CT-GAN \cite{ying2019x2ct:8}                                       &  \underline{0.280} &  \underline{18.13} &  \underline{0.072}  &  \underline{0.163}
             \\
& CCX-rayNet \cite{ratul2021ccx} &  0.122 &  16.62 &  0.093  &  0.216
             \\
& Ours  &  \bf{0.315}  &  \bf{19.85}  &  \bf{0.047}  &  \bf{0.126}  
             \\
    \end{tabular}}
\end{center}
\vspace{-10pt}
\caption{Quantitative results on both in-distribution and several out-of-distribution datasets using only two input X-rays. We used our model weights for 5,000 training optimization steps, trained with Swin-U-Net. Other models weights are from 100 epochs ($\sim$ 90k iterations).}
\label{tab:comparisons}
\vspace{-10pt}
\end{table}

\paragraph{Mapping reformulation} In Figure \ref{fig:baseline_experiments}, we present results from initial experiments comparing an asymmetrical 2D to 3D strategy versus our reformulation as a 3D to 3D mapping. The 2D to 3D approach involves aggregating the individual encoded 2D view features, which serve as input to a 3D decoder. However, this results in highly blurred outputs due to the information content loss from the latent encoding. Despite the simplicitiy of such an approach in terms of feature alignment, important fine-grained details are missing. In contrast, our approach reformulates the problem into a 3D to 3D mapping in a simple architecture, achieving high-fidelity image translations even when using a single input view. We ablate our 'repeat and concatenate' preprocessing pipeline by instead concatenating noise sampled from a normal distribution. When the vector $\vec{z}$ is fixed for each patient, the model causes overfitting, while gathering a new $\vec{z}$ for each iteration results in mode collapsing, where it produces the same output irrespective of the input.

\paragraph{Ablations \& comparisons} Our approach allows the use of multiple views as input without modifying our model's architecture. We observe that the model benefits from additional views when tested on in-distribution inputs (Fig. \ref{fig:views_ablation}). However, this correlation was found not strictly hold for out-of-distribution inputs. This suggests that when there are disparities in view alignment or style-domain features compared to the training distribution, the model predominantly relies on views that closely resemble the learned patterns. Investigating whether certain projections contain more valuable information than others is a valuable research direction for future work. Overall, we find that the combination of 2 or 4 views to be empirically effective (Table \ref{tab:views_ablations}).

\begin{figure*}[!ht]
    \centering
    \includegraphics[width=1.0\linewidth]{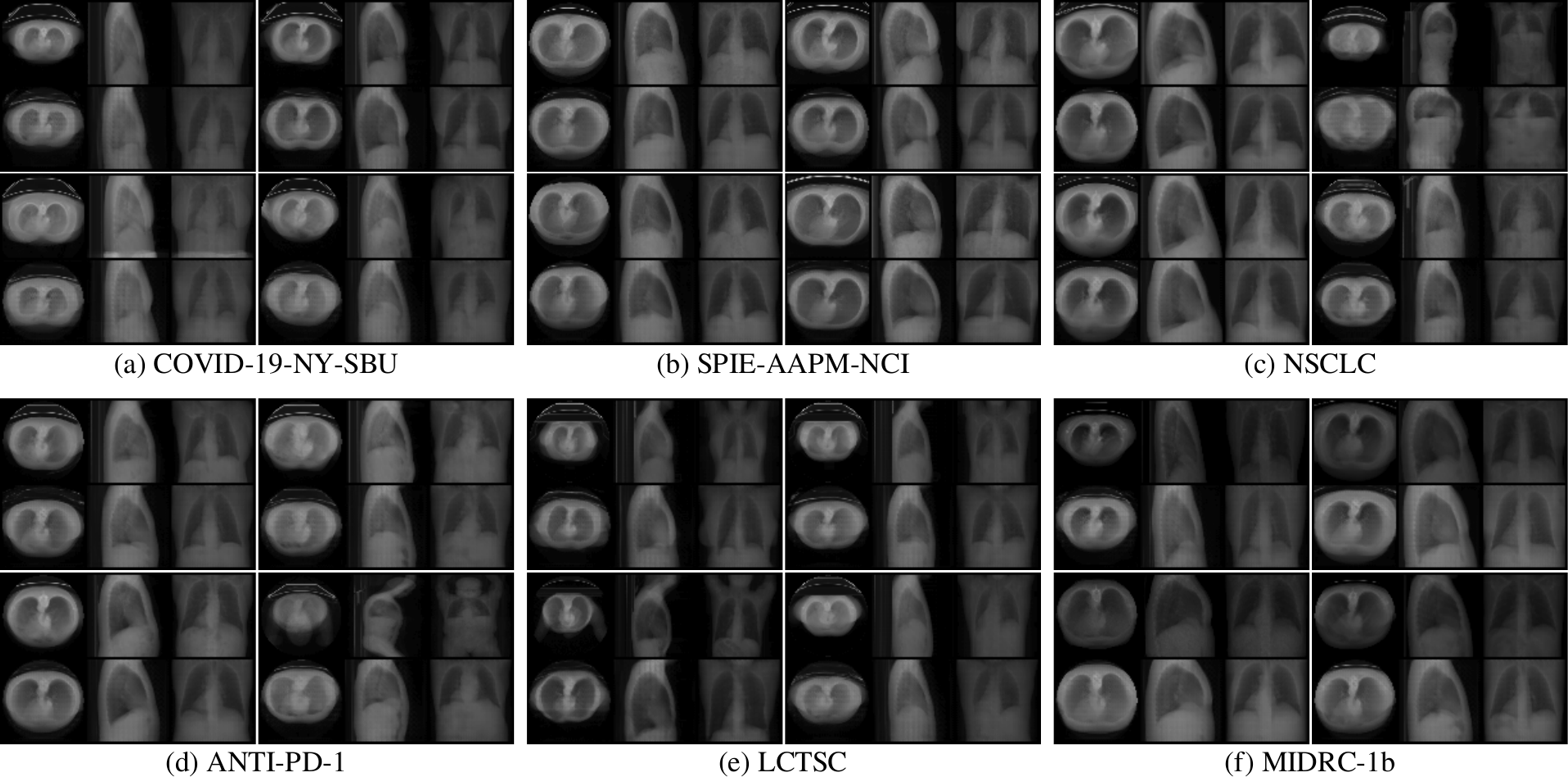}
    \caption{CT projections from generated 3D volumes on various out-of-distribution lung datasets. Model weights were selected from iteration 5,000 on the LIDC-IDRI dataset \cite{armato2011:lidc}. GT projections are displayed in odd rows, while our model's outputs are shown in even rows.}
    \label{fig:out-dist-qualitative}
\end{figure*}

\begin{figure}[!ht]
    \centering
    \includegraphics[width=1.0\linewidth]{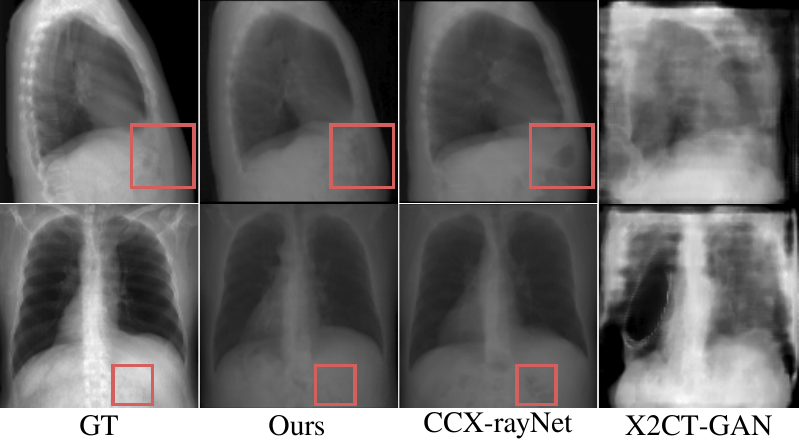}
    \caption{Examples of the correlated 3D projections from our proposed approach compared to alternative supervised methods X2CT-GAN \cite{ying2019x2ct:8} and CCX-rayNet \cite{ratul2021ccx}.}
    \label{fig:high-res-comp}
    \vspace{-15pt}
\end{figure}

In Table \ref{tab:comparisons} we compare our approach to paired alternative methods in terms of quality of the 3D outputs for both in- and out-of-distribution inputs. Despite variations in datasets originating from different imaging systems, resolutions, and patients' health conditions, our model demonstrates superior performance across all metrics and maintains consistency across all out-of-distribution datasets. Refer to Figures \ref{fig:out-dist-qualitative} and \ref{fig:high-res-comp} for qualitative results.

\begin{figure}[ht]
    \centering
    
    \begin{minipage}[b]{\columnwidth} 
        \centering
        \fontsize{8pt}{9pt}\selectfont
        \begin{tabular}{lcccc}
        \multicolumn{5}{c}{(a) In-of-distrib. inputs (LIDC-IDRI dataset \cite{armato2011:lidc})}
                \\ \midrule[0.1pt]
        Oblique views ablation &  $\uparrow$  SSIM  &  $\uparrow$ PSNR  &  $\downarrow$ MSE  &  $\downarrow$ MAE    
             \\    \midrule[0.1pt] 
        no alignment     &  \bf{0.5001}  &  \bf{24.0513} &  \bf{0.0178}  &  \bf{0.0682}
                 \\
        coarse alignment     &  0.4248  &  22.5182 &  0.0259  &  0.0872 
                 \\
        \\
        \multicolumn{5}{c}{(b) Out-of-distrib. inputs (COVID-19-NY-SBU dataset \cite{saltz2021stony:covid-19})}
                \\ \midrule[0.1pt]
        no alignment     &  \bf{0.3581}  &  \bf{21.9844} &  \bf{0.0272}  &  \bf{0.0881}
                 \\
        coarse alignment     &  0.3365  &  21.0851 &  0.0340  &  0.1026                        
                 \\
        \end{tabular}
        \caption{Results on oblique views with coarse alignment.}
        \vspace{5pt}
        \label{fig:oblique_views}
    \end{minipage}
    \nextfloat
    
    \begin{minipage}[b]{\columnwidth} 
        \centering
        \begin{subfigure}[b]{0.49\linewidth}
            \centering
            \includegraphics[width=\linewidth]{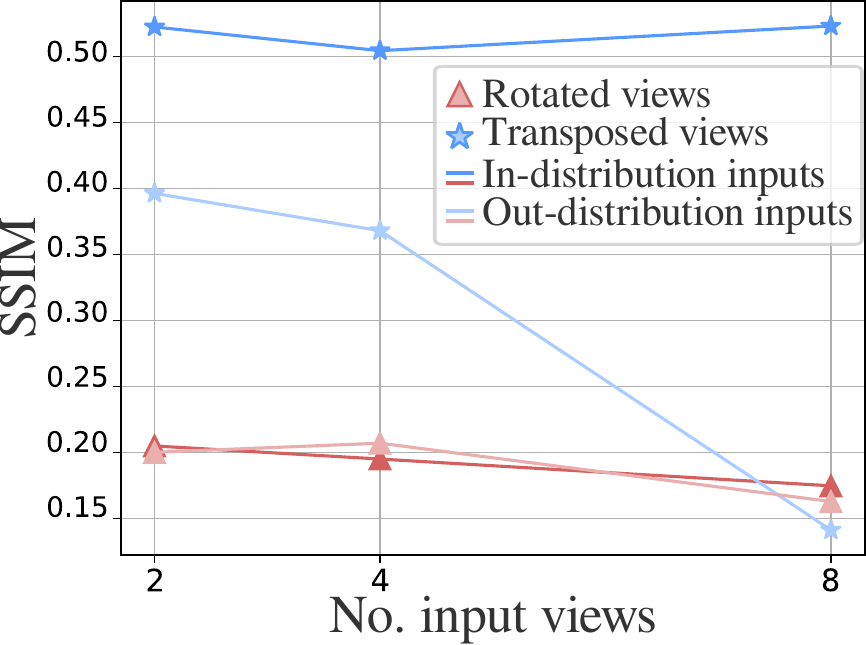}
        \end{subfigure}
        \hfill
        \begin{subfigure}[b]{0.49\linewidth}
            \centering
            \includegraphics[width=\linewidth]{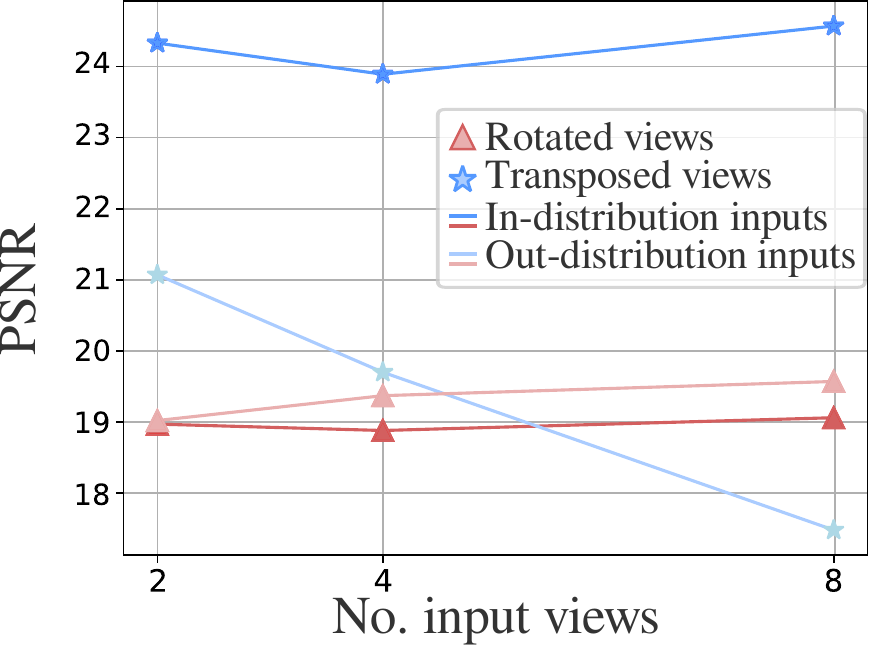}
        \end{subfigure}
        \caption{Comparison between only transposing perpendicular views (indicated by a star marker) and a coarse alignment (triangle marker), using two, four, and eight input views. Smoother colors represent results for out-of-distrib. inputs (MIDRC dataset \cite{tsai2020:midrc}).}
        \label{fig:alignment_comp}
    \end{minipage}
    \vspace{-25pt}
\end{figure}

\begin{figure*}[!ht]
    \centering
    \includegraphics[width=1.0\linewidth]{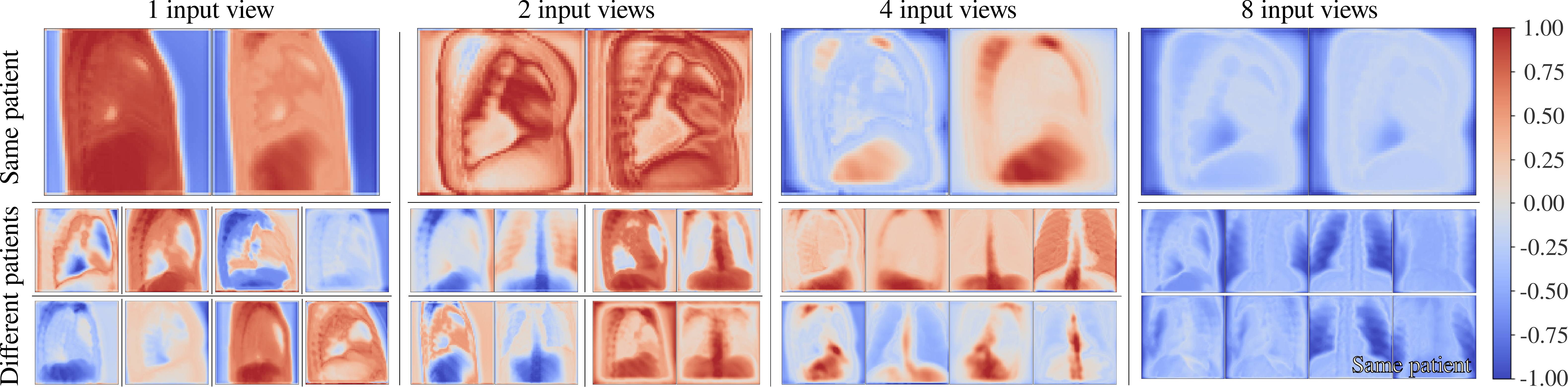}
    \caption{Gradient magnitudes with respect to the input views (X-rays) of the proposed image translation model. The \textit{upper row} displays the variability within slices extracted from the 3D input volume of the same patient, while the \textit{bottom row} shows the attribution among slices from different patients.}
    \label{fig:grads_views}
\end{figure*}
\paragraph{Gradients \& patient-specific learning} In Figure \ref{fig:grads_views}, we visualize the magnitudes of the gradients of our model with respect to a varying number of input views. We observe higher variability across different patients, indicated by the color variations for one, two, four, and eight views independently. On the other hand, if we analyze intra-patient gradients, we notice lower variability, suggesting consistent patterns captured in views coming from the same patient. With our approach, the generative model implicitly learns patient-specific representations, where outputs reflect differences in anatomical structures and/or pathological conditions present in the input data from each patient.

\paragraph{View alignment} We study the effect of aligning the input views during the preprocessing stage prior to the repeat and concatenation operations. Initially, we experimented without any alignment, which resulted in outputs containing checkerboard-like artifacts that diminished after an additional 1k training iterations. However, a simple transposition of perpendicular views from coronal and sagittal planes alleviated this. To test our model capacity on views outside of such geometry, we tested on four oblique projections, for which we find that our model performs equally well without relying on any sort of alignment (Fig. \ref{fig:oblique_views}). We also investigated a coarse alignment by approximating their locations through 3D rotations using data transformations from Kornia's library (Figs. \ref{fig:oblique_views}, \ref{fig:alignment_comp}). However, we observe that this might required a precise and potentially non-affine transformation which is non-trivial. Overall, we find that the model learns some invariance to this, where a simple transpose consistently yields improved results without apparent artifacts (Fig. \ref{fig:alignment_comp}). While precise projection alignment in 3D space could potentially enhance results, we consider there may be a tradeoff between alignment robustness (e.g., due to patient movement) and generation accuracy. In the future, an iterative 3D alignment approach would be worthwhile investigating as a secondary objective.

\subsection{Implementation Details}
Our training algorithm is based on the neural optimal transport approach by Korotin \textit{et al.} \cite{korotin2022neural}. In this approach, the mapping network $g_\varphi$ undergoes $k=10$ iterations while the parameters of the potential network $d_\phi$ remain frozen. Then, a single training step is performed for $d$, unfreezing its parameters while freezing those of $g$. Our feature extractor $f$ is parameterized by a 3D convolutional neural network with Kaiming weight initialization. We set $\lambda = 0.1$ to control the degree of regularization of the reconstructed outputs by $d$. Improving generation quality could be achieved by scaling to higher resolutions and adjusting the $\lambda$ parameter. Our networks are trained using the AdamW optimizer with parameters $\beta_1=0.5$, $\beta_2=0.999$, $\epsilon=0.001$, and a learning rate of $10^{-5}$ with weight decay. We apply a combination of differential augmentations \cite{zhao2020differentiable}, including random contrast, rotation, and horizontal flip, to the input 2D X-rays in all experiments. 3D reconstructions are saved in both numpy and .mha formats for visualization using medical image software such as 3D Slicer.
\section{Limitations and future directions}
The approach, while simple and intuitive, has several clear limitations. Firstly, our current architecture attempts the non-linear transformation in a `single' transformative step, resulting in uncertainty and therefore blurriness. We expect improved performance through an iterative multi-step transformation, such as a modern probabilistic diffusion generative model. Secondly, while we found some invariance to the input alignment, we would like to investigate whether the concatenated 3D inputs could be better aligned to match the target 3D locations and therefore support better modeling of some high-frequency details near their corresponding specific 3D slices. For example we could potentially optimize the input affine transformations, in particular their roto-translations, according to a secondary objective, repeated at inference as part of the modeling.

\section{Conclusion} \label{s:conclusion}
In conclusion, we found that simply repeating the 2D inputs into concatenated 3D volumes, then treating the 2D to 3D translation problem as a 3D to 3D translation task, leads to improved correlation between the synthesized outputs over more sophisticated multi-view latent integration approaches. While we expect other off-the-shelf 3D to 3D conditional generative modeling approaches to be immediately applicable within our framework, we found particular success by directly applying 3D to 3D neural optimal transport to attain highly correlated 3D synthesized outputs with their corresponding 2D inputs.
The overall proposed approach is fast to train, data efficient, stable, and generalizes well to new out-of-distribution views making it applicable in a real-world clinical setting. However it does exhibit blurriness where there is uncertainty in the outputs; in the future it would be worth investigating iterative alignment optimization for the repeated 3D inputs as a secondary objective to the generative modeling, repeated during inference, potentially mitigating this uncertainty. 

\vspace{15pt}
\paragraph{Acknowledgments} This work was supported by CONAHCyT, and Durham University.


{
    \small
    \bibliographystyle{ieeenat_fullname}
    \bibliography{main}
}


\end{document}